%% file: ICRA2021_main.tex
\documentclass[conference]{IEEEtran}
\IEEEoverridecommandlockouts
\usepackage{cite}
\usepackage{amsmath,amssymb,amsfonts}
\usepackage{algorithmic}
\usepackage{graphicx}
\usepackage{textcomp}
\usepackage{lipsum}  
\usepackage{xcolor}
\usepackage{hyperref}

\begin{document}

\title{\LARGE \bf Effects of Interfaces on Human-Robot Trust:\\Specifying and Visualizing Physical Zones\\
\thanks{Funded in part by NSF grants NRI 2024872 and 2024673}
}
\author{Marisa Hudspeth$^2$, Sogol Balali$^1$, Cindy Grimm$^1$, Ross T. Sowell$^2$\\}


\maketitle

\begin{abstract}
In this paper we investigate the influence interfaces and feedback have on human-robot trust levels when operating in a shared physical space. The task we use is specifying a ``no-go'' region for a robot in an indoor environment. We evaluate three styles of interface (physical, AR, and map-based) and four feedback mechanisms (no feedback, robot drives around the space, an AR ``fence'', and the region marked on the map). Our evaluation looks at both usability and trust. Specifically, if the participant trusts that the robot ``knows'' where the no-go region is and their confidence in the robot's ability to avoid that region. We use both self-reported and indirect measures of trust and usability. Our key findings are: 1) interfaces and feedback do influence levels of trust; 2) the participants largely preferred a mixed interface-feedback pair, where the modality for the interface differed from the feedback. 

\end{abstract}

\begin{IEEEkeywords}
Interfaces, Trust, Safety, Usability
\end{IEEEkeywords}

\input{Text/1.introduction.tex}

\input{Text/2.related_work.tex}
\input{Text/3.method.tex}
\input{Text/4.results.tex}
\input{Text/5.discussion.tex}
\input{Text/7.conclusions.tex}

\input{Text/9.appendix}


\bibliographystyle{IEEEtran}
\bibliography{references.bib}


\end{document}

%% file: Text/1.introduction.tex
\section{Introduction}

Robots are entering both public and private spaces and interacting with the broader public. Unlike digital devices such as phones and tablets, the robot moves around, and interacts with, the physical environment. We are broadly interested in how to design interfaces for the general public that enable spatial communication between the human and the robot --- e.g., put that there, don't go there, go down this path. Obviously, an interface should be {\em usable} (and effective). However, we also argue that --- because people naturally assign agency to robots --- an interface should also engender an appropriate level of {\em trust} that the robot both understands the instruction(s) and is capable of following them. In this paper we investigate both usability and trust in the context of specifying a ``no-go'' region in an indoor environment.

The spatial human-robot task we focus on in this paper is specifying a region the robot is {\bf not} suppose to enter (the no-go region). There are several possible contexts for this task, ranging from safety (for the robot, humans, or objects) to privacy (don't go in the bathroom); we use the context of ``breakable objects''. This enables an {\em indirect} measure of human-robot trust (how willing the participant is to put breakable objects in the ``no go'' space) in addition to the standard self-reported trust measures. Note that --- from an implementation stand-point --- this is relatively easy to implement by simply modifying the robot's map.

We separate the robot-human interaction into two parts, the human communicating to the robot (the interface) and the robot communicating {\em back} to the human (the feedback). Specifically, the task itself naturally lends itself to three interface types which move from indirect (specifying points on a map) to semi-direct (an AR interface where the participant clicks on a map) to direct (putting physical cones around the region). Similarly, we compare indirect feedback (shaded region on the map), to semi-direct (a virtual fence in AR), to direct (the robot drives around the region). We also include a {\em no} feedback option to determine how much of a role feedback plays on trust. 

We ran a user study to evaluate both the usefulness and trust levels of our interfaces and feedback. We use both self-reported measures of trust and usefulness as well as novel indirect measures of trust (where participants placed breakable objects and the size of their no-go regions). Our study is primarily a between-subjects design with each participant experiencing each interface, but on slightly different (but related) tasks. 

Specifically, our study hypotheses are:
\begin{itemize}
    \item[H1] The physical interface will engender more trust than the map one.
    \item[H2] Physical feedback will engender more trust than the map feedback, and any feedback more than no feedback.
    \item[H3] The AR interface will be preferred over the map one.
    \item[H4] The physical feedback will be preferred over the map one. 
\end{itemize}

%% file: Text/2.related_work.tex
\section{Related Work}
\label{sec:related}
The concept of trust has been studied in many contexts and fields.  It is a multifaceted concept that encompasses many relational components including prior expectations, confidence level, existing risk or uncertainty, and level of reliance~\cite{Billings2012b}.  There are over 300 documented definitions of trust across disciplines, and a general lack of consensus on what it is, and how to measure it~\cite{Adams2003,Schaefer2013}. In this section, we give a brief overview of the related work in human-robot interaction (HRI) and specifically on the trust measures that have been developed.


 
In HRI, studies of trust have considered several factors such as the robot’s perceived competence, benevolence, predictability and transparency~\cite{Gulati2018,Boyce2015} as well as the human's self-efficacy and perceived ability to use and interact with a robot.
Recent studies have applied meta-analytic methods to the available literature on trust and HRI to assess and quantify the effects of human, robot, and environmental factors on perceived trust in HRI, summarize the relationships between the trust variables~\cite{Hancock2011Meta}, and introduce certain categorization of trust in HRI \cite{LAW2021}. Other studies emphasized the importance of being able to measure trust in HRI and developed scales to measure it ~\cite{Schaefer2016Measure,Charalambous2016,Yagoda2012YouWM,Underhaug2019}. 


Studies have explored the effects of a robot's functional capabilities (e.g., behavior, reliability, and accuracy), while other studies focused on evaluating the impact of robot features (e.g., level of automation, anthropomorphism, robot type, personality, and intelligence) on trust. Studies of human-robot systems have focused on human traits~\cite{Kidd2011,Scopelliti2003}, as they affect trust development and human state (e.g., stress, fatigue, workload) and its role on trust development. Other than robot and human related determinants, the environment also influences trust in HRI~\cite{Schaefer2013}. 
Thus, many studies have explored environment related factors (i.e., task related factors, team related factors) that impact trust, or they have specifically studied how trust is impacted in particular environments~\cite{Gabrecht2016,Charalambous2016,Pinxteren2019,Schaefer2015AutonomousVehicale}. 



To evaluate trust in HRI, a variety of subjective self-reported scales in the form of questionnaires have been developed ~\cite{Donnellan2006MiniIPIP,Nomura2004Nars,Bartneck2009,Yagoda2012YouWM,Ullman2019MeasuringGA}. Schaefer~\cite{Schaefer2016Measure} combined several of these scales and developed the Trust Perception Scale-HRI that focuses on measurable factors of trust specific to the human, robot, and environmental elements.
Chita-Tegmark et al. evaluated this scale, along with two other trust questionnaires to refine measures of trust so that they are exploratory and generalizable with respect to social dimensions of trust ~\cite{Tegmark2021}. Objective measures of trust, although not as commonly used as the subjective measures, have been employed to measure trust in several studies and are recently classified into categories, namely, behavioral change, task intervention, following advice, and task delegation \cite{LAW2021}. While the findings that emerge from trust measurements are of great value, several studies have focused their attention on determining whether these findings could be reproducible when conducting replications across distinct robots and/or modified study design to ensure validity and generalizability of the findings \cite{Ullman2021,Law2021ATC}.


The work reported in this paper contributes to the study of trust by 
evaluating human trust 
that a robot will respect a no-go region by not entering it. To do that we asked people to either position a table with a fragile item on it inside the no-go region or to place items with different levels of fragility on tables that are placed inside the no-go region. We introduce three interfaces (Physical, Map, AR) for specifying a no-go region and four feedback options (Physical, Map, AR, Control).  This allows us to assess the extent to which people's trust gets impacted by different interface and feedback options.


 

%% file: Text/3.method.tex
\section{Methods}
\label{sec:methods}

The aim of this study is to examine how human trust is influenced by different robotic interfaces and feedback conditions. Our specific task is marking off a region of a room as a ``no-go'' region. Our trust measures are specific to the task: Does the robot know about the no-go region? Will it respect it? We measure trust both directly (self-report questionnaires) and indirectly (by how they arrange items in the no-go area).

Our three interfaces move from direct (physically placing cones in the environment) to indirect (clicking on a map), and are broadly modeled after the types developed by Reuben~\cite{Rueben2016} for physical privacy specification. Our three feedback conditions similarly move from direct (robot drives around the no-go region) to indirect (marks on a map). Our forth feedback condition is no feedback. We introduce each interface in a random order over three tasks which move from very structured to free-form. To reduce the total number of participants required (and reduce participant confusion) we match each interface with either its corresponding feedback condition OR no feedback (i.e., we do not try all possible interface-feedback combinations). In the final step of the procedure the participants repeat the third task but are given the option to choose the interface/feedback combination. Our analysis is primarily a between subjects one, comparing the three different interface plus matching feedback to the interface without feedback. 

We next describe the interfaces and feedback options used in the study (Section~\ref{sec:interfaces}), the three tasks (Section~\ref{sec:tasks}), the surveys used (Section~\ref{sec:surveys}), the full study procedure (Section~\ref{sec:procedure}),  participant demographics (Section~\ref{sec:participants}), and finally, how we measured trust (Section~\ref{sec:measures}).

\subsection{Interfaces and Feedback Options}
\label{sec:interfaces}
Our no-go region is defined to be a quadrilateral. We developed three interfaces (top row, Figure~\ref{fig:specification}) for specifying the four corners of the no-go region: 
 \begin{enumerate}
    \item \textbf{Physical}: The user places four plastic cones.
    \item \textbf{Map}: The user clicks four points on a map of the room.
    \item \textbf{AR}: The user places four virtual cones by tapping through the smartphone while looking through it.
 \end{enumerate}
 \input{Text/InterfacePictures.tex}
 
 We define three different feedback conditions (bottom row, Figure~\ref{fig:specification}) plus a control condition (no feedback).  
 \begin{enumerate}
    \item \textbf{Physical}: The robot drives around the room, avoiding the no-go region. This takes about 12 seconds. 
    \item \textbf{Map}: The no-go region is shaded red on the map of the room.
    \item \textbf{AR}: A virtual fence appears on the smartphone's screen.
    \item \textbf{Control}:  No feedback is provided.
 \end{enumerate}
\noindent For the first three tasks each interface was either paired with its corresponding feedback condition or no feedback was provided. In the last task the participants were allowed to mix and match the interface and feedback types.  For example, specifying a no-go region using the AR interface and then observing the physical feedback. 
 
To ensure the interfaces were comparable, we ran a pilot study to verify that the ease of use (as measured by a survey) and time to specify a region were similar (within 1/10 of the total time). We also confirmed that when attempting to mark the same region with each interface, the resulting regions were roughly the same size (+- 4\% area).

During the study, the study coordinator explicitly told the participants that the robot had access to the map/cone location.

\subsection{Tasks}
\label{sec:tasks}
The experiment consists of three tasks. 

\textbf{Task 1 (Placing a Table):} The study coordinator created a no-go region and then presented the participant with a rickety table that had a fragile object on top of it. The participant was told that if the robot were to bump into the table, the object might fall and break. The participant was asked to place the table inside the no-go region where they thought it would be safe from the robot. 

\textbf{Task 2 (Placing Objects):} The study coordinator created a no-go region and placed two identical tables inside it, one close to the edge of the region and one far from the edge of the region. The participant was presented with four items (two durable and two fragile) and was asked to place them on the tables so that they would be safe from being knocked off by the robot (see Figure~\ref{fig:items}).
\input{Text/task2Items}

\textbf{Task 3, Part 1 (Creating a Region):} The participant was asked to create a no-go region around a rickety table with a fragile object on top of it. For the first part they were assigned an interface plus feedback (or no feedback) combination.

\textbf{Task 3, Part 2 (Choosing an Interface):} Task 3 was repeated a second time, but in this instance participants were allowed to choose whichever interface and feedback combination they wanted to use. 

\subsection{Surveys}
\label{sec:surveys}
    This study contains three surveys described below.  A complete list of survey questions and summary data are publicly available\footnote{\label{note1}Survey questions and summary data: \href{http://cs.rhodes.edu/~sowellr/research/trust/survey.pdf}{Link}}.

The \textbf{entry survey} consists of standard demographic questions (e.g., age, gender, education level) and questions that evaluated participants' initial attitude toward robots using the NARS scale ~\cite{Nomura2004Nars}. Lastly, we evaluated participants' information processing style, computer self-efficacy, attitude toward risk, and willingness to explore or tinker using the GenderMag toolkit ~\cite{Burnett2016} .

The \textbf{between task survey} consists of 3 questions related to trust (recognize the no-go region, know to avoid it, and actually avoid it in practice). These were all 5-pt Likert scales ranging from ``Never'' to ``Always.'' Participants were also asked to specify how differently they would answer this question if they were (or were not) provided with feedback (5-pt Likert scale from ``None at all'' to ``A great deal'').

The \textbf{usability survey} asked participants to rate the robot's overall competence, if it was capable of avoiding the region, and if it would choose to enter the no-go region even if it was capable of avoiding it. They were also asked to rate the ease of use and usefulness of each interface. These are all 5-pt Likert scales. This survey was given after Task 3 part 1.

\subsection{Procedure}
\label{sec:procedure}
Participants started the study with the entry survey. They then performed each of the three tasks in turn, each time with a different interface/feedback combination. After the third task (part 1) they took the usability survey.  Participants then repeated the third task, but this time they chose one of the interfaces and one of the feedback options. They were not required to match the interface and feedback type (and many did not); the no feedback condition was not given as an option.

The order of the interface/feedback pairs was randomized, and each interface was further randomly assigned to have either the corresponding feedback or no feedback. This results in a total of 6 conditions. A Latin squares design was used to ensure that the order of the interfaces and the feedback/no feedback conditions were evenly balanced (six conditions total, five participants per each condition, for a total of 30 participants). 

After each task, if the participant saw feedback, they were asked if {\em not} seeing the feedback would change their answer. If the participant did {\em not} see the feedback, they were shown the matching feedback and asked if seeing the feedback would have changed their answers. This provided additional information on the usefulness of the feedback and also ensured that all participants saw all feedback options before completing the second part of task three.

 \input{Text/ResultsFigures/H1H2}

\subsection{Participants}
\label{sec:participants}
Participants (30 total) consisted of Rhodes College faculty (16.7\%), staff (3.3\%), and students (80\%). 80\% of participants were 18-24 years old, with the remaining participants ranging from 36-60 years old. 30\% identified as women and 70\% as men.

\subsection{Measures}
\label{sec:measures}

\noindent {\bf Task 1:} The distance from the table to the edge of the no-go region (shorter indicates more trust). 

\noindent {\bf Task 2:} For each object, placement on the close versus the far table, and distance from the edge of the table (shorter indicates more trust).

\noindent {\bf Task 3, part 1:} The distance of the table to the edge of the no-go region (shorter indicates more trust). 

\noindent {\bf Task 3, part 2:} The area of the no-go regions (smaller indicates more trust) and the placement of the table in the no-go region.

\noindent {\bf Survey measures:} The 5-pt Likert survey scales were converted to a score between 1 and 5. 

%% file: Text/InterfacePictures.tex
\begin{figure}[ht]
\centering
\includegraphics[width=\linewidth]{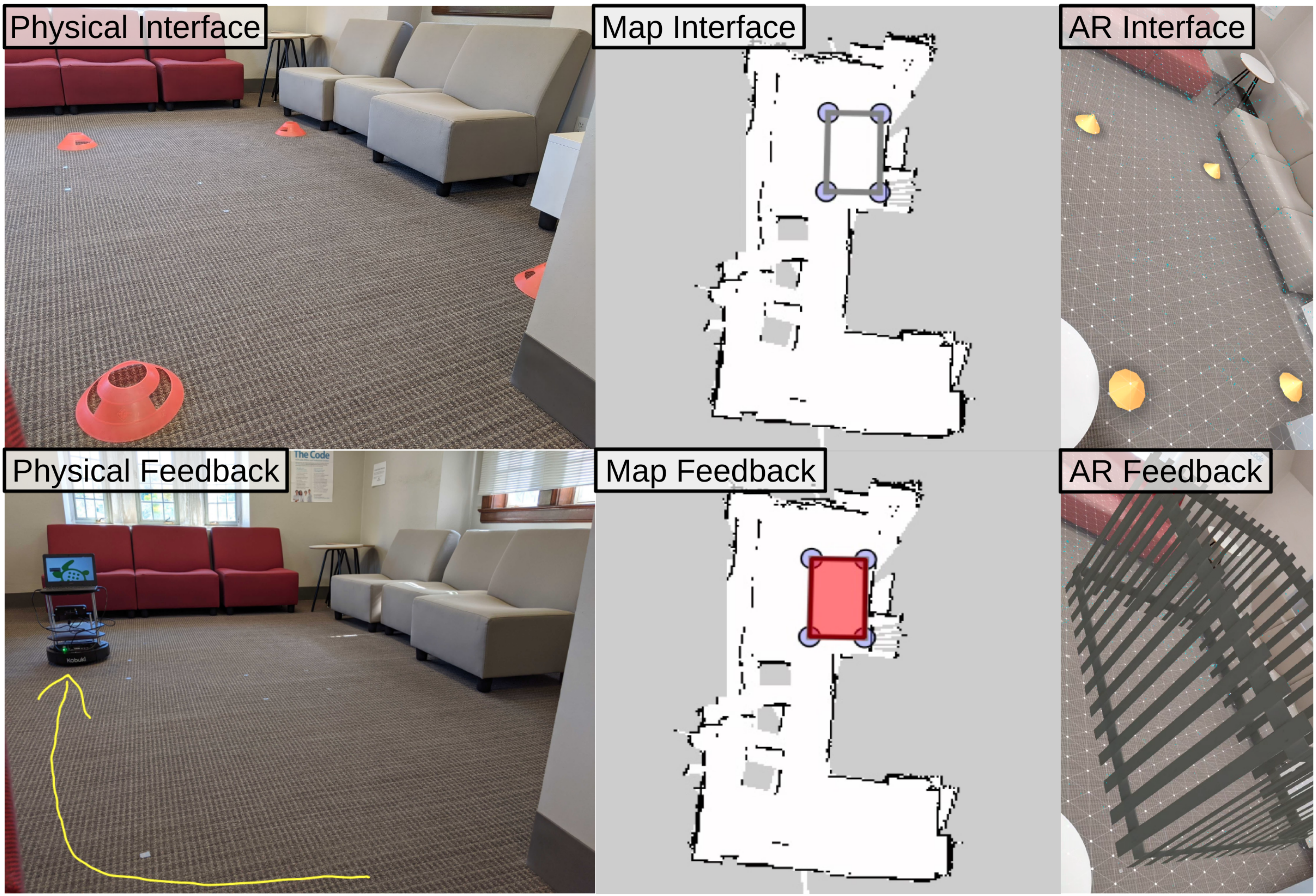}
\centering
\caption{Interfaces and feedback options for specifying the no-go region.}
\label{fig:specification}
\end{figure}

%% file: Text/task2Items.tex
\begin{figure}[ht]
\centering
\includegraphics[width=\linewidth]{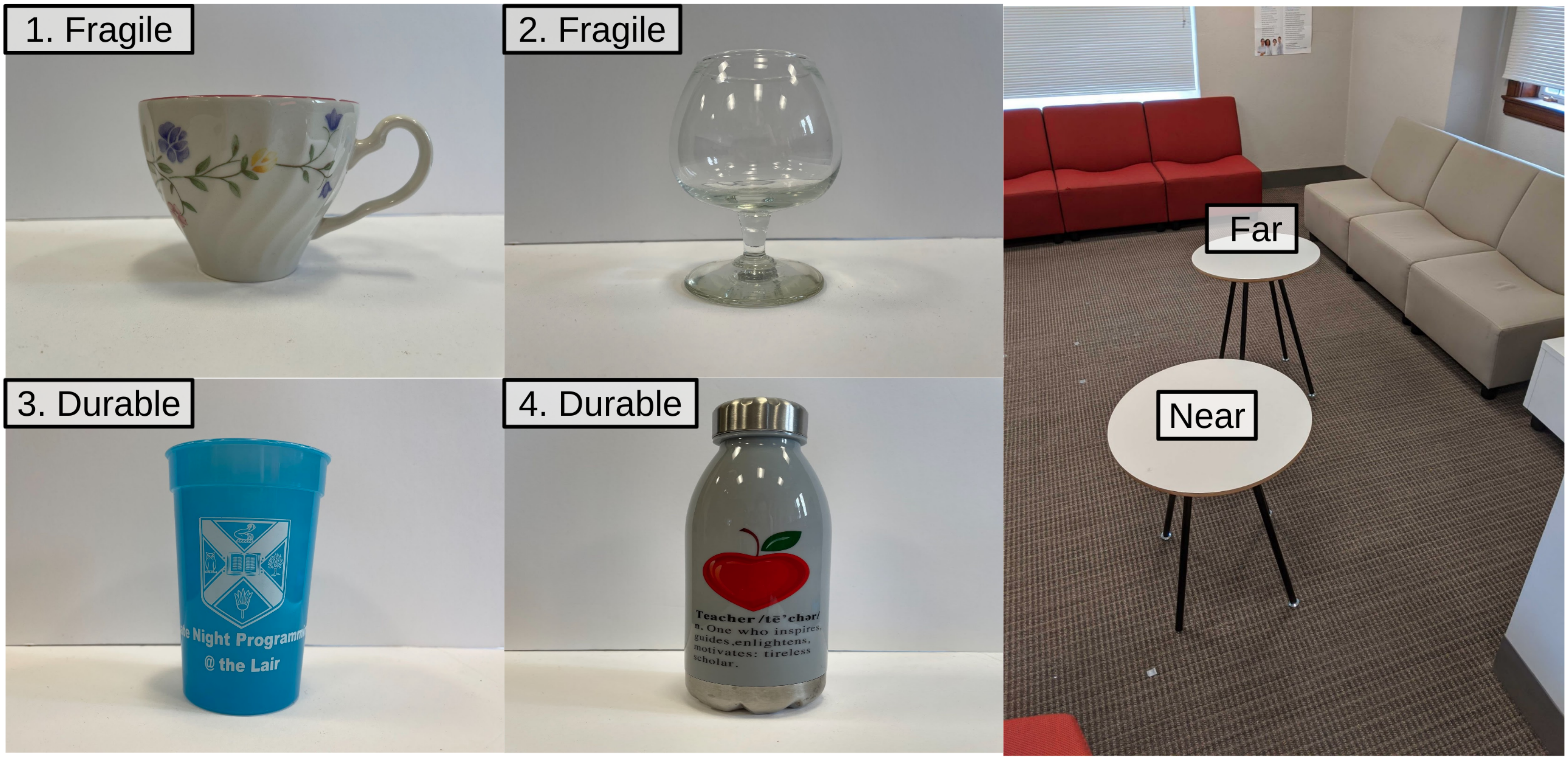}
\caption{Objects and tables for Task 2.}

\label{fig:items}
\end{figure}

%% file: Text/ResultsFigures/H1H2.tex
\begin{figure*}
\begin{minipage}{0.3\textwidth}
\small
 \begin{tabular}{|l|ccc|}
 \hline 
 Comparison       &  Supp & Eq & Not \\ 
 \hline 
 Phy I $>$ Map I & 5 & 9 & 0  \\
 AR I       $>$ Map I & 10 & 4 & 0 \\
 Phy I $>$ AR I  & 2 & 8 & 4 \\
 \hline
 Phy IF $>$ Map IF & 9 & 2 & 3 \\
 AR IF       $>$ Map IF & 11 & 3 & 0 \\
 Phy IF $>$ AR IF & 4 & 4 & 6 \\
 \hline
 Phy F $>$ No F & 4 & 4 & 6 \\
 AR F       $>$ No F & 7 & 5 & 2 \\
 Map F $>$ No F & 3 & 1 & 9 \\
 \hline
Feedback $>$ No F & 1 & 7 & 4 \\
\hline
 \end{tabular}
\end{minipage}
\begin{minipage}{0.65\textwidth}
    \includegraphics[height=0.375\linewidth]{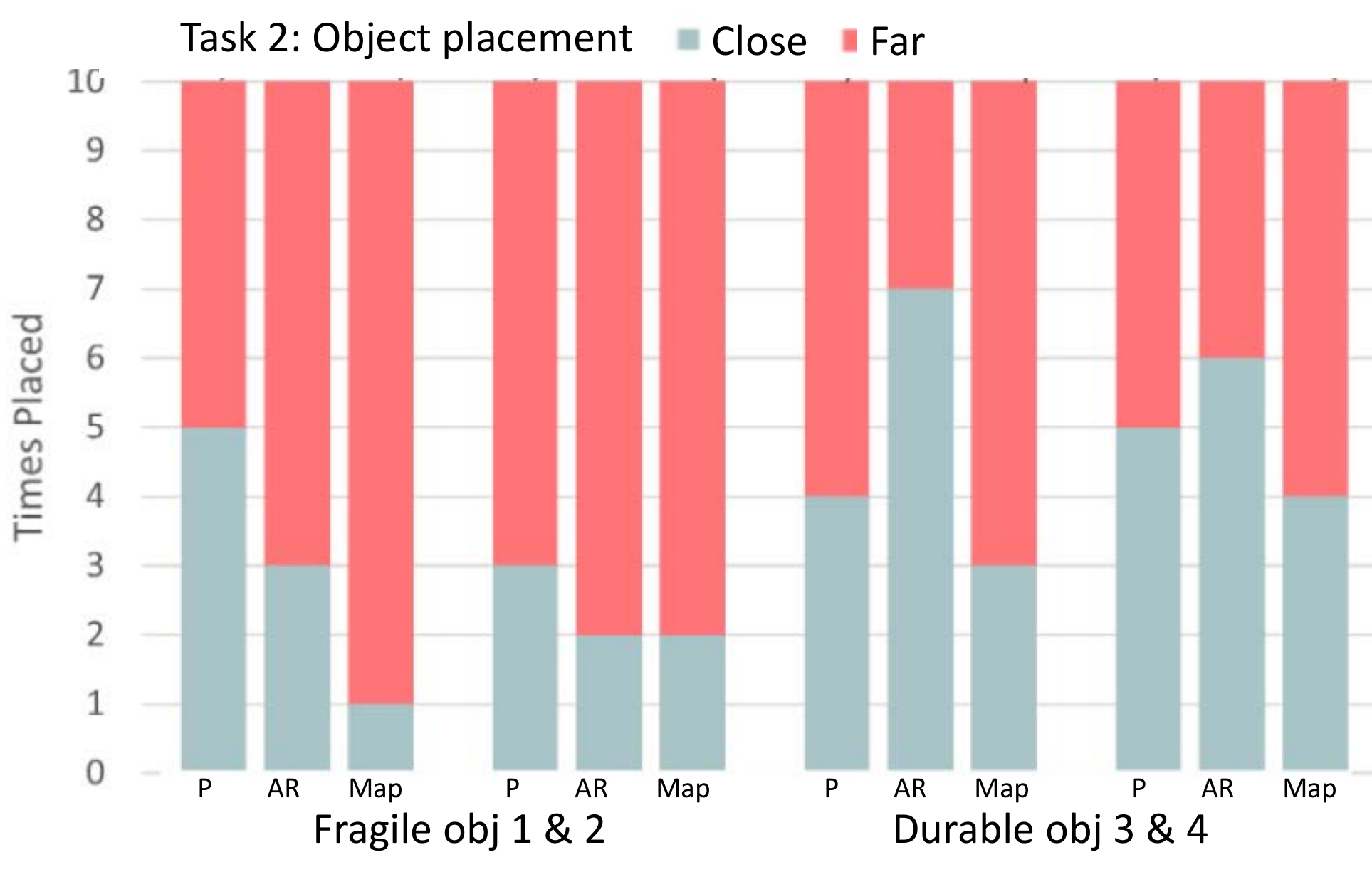}
    \includegraphics[height=0.375\linewidth]{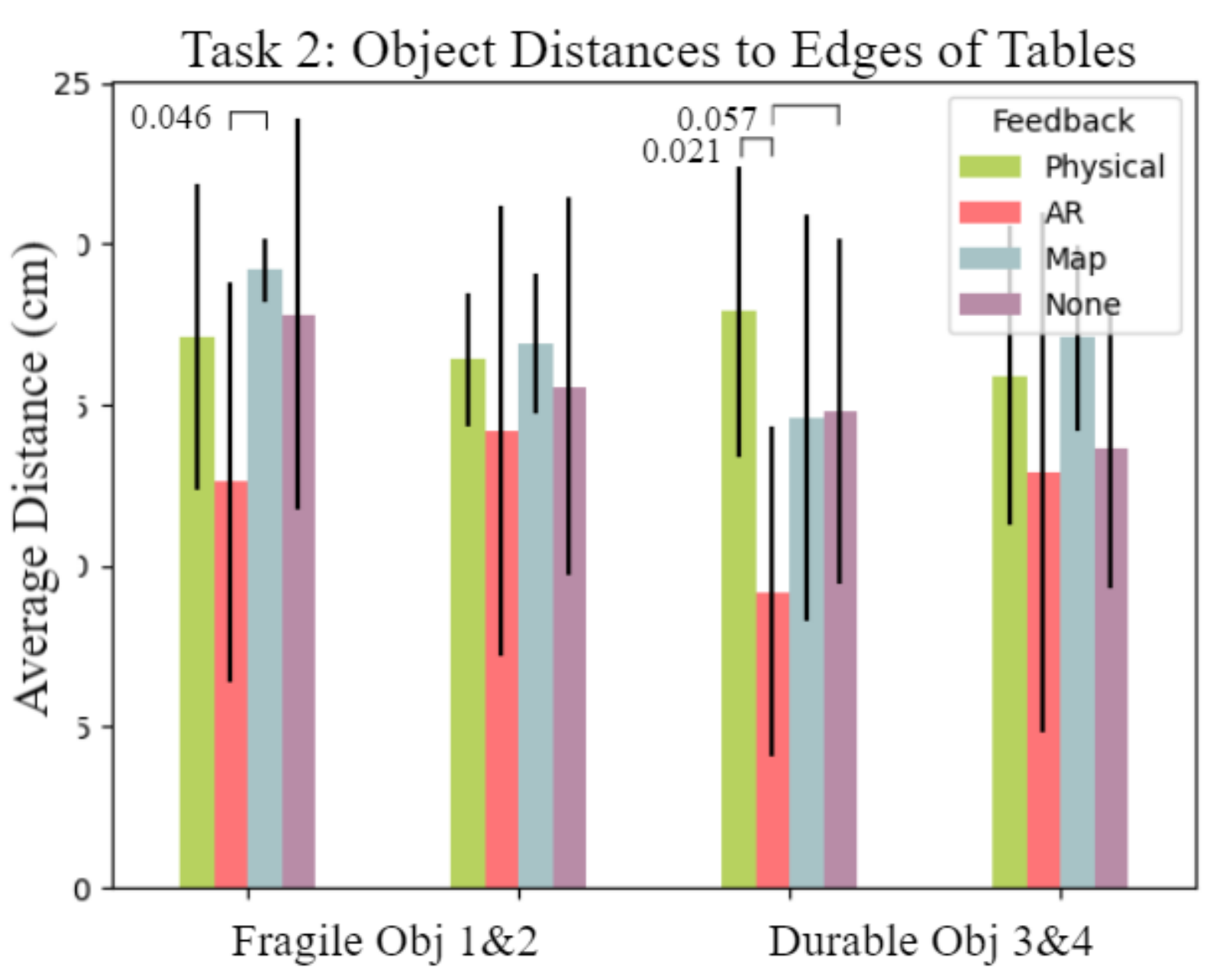}
\end{minipage}
 \caption{Measures that support, are equivalent, or not support trust \textbf{(Left)}. From top to bottom: Interface (any feedback), Interface plus paired feedback, Interface with versus without feedback, any interface, with and without feedback. Object placement on close or far tables \textbf{(Middle)}. Object distance to edge of table \textbf{(Right)} .}
 \label{tab:H1AndH2}
\end{figure*}

%% file: Text/4.results.tex
\section{Results}

We next present results for the hypotheses; we combined 1 with 2 and 3 with 4. Complete summary data are publicly available (see footnote~\ref{note1}). We summarize the numerical results here.

\subsection{H1 \& H2: Interface/Feedback and Trust}
\label{sec:res_H1}

\input{Text/ResultsFigures/H3}

Each interface was seen by each participant in one of the first three tasks. We combine self-reported trust measures (survey each task, 3 total) with indirect trust measures (object or table placement, area of region, 11 total) and use both between and within-subject data.  Table~\ref{tab:H1AndH2} (left) summarizes the number of indirect and survey measures that support H1 (interface/feedback) and H2 (feedback versus no feedback). The top 3 rows compare the interface trust measures regardless of whether or not the participant saw the feedback; all 30 participants saw all conditions, just with different tasks. We separate out this data into participants who saw the feedback versus not (15 participants each). In the final row we compare just the feedback question (within subjects only).  Full details are in Appendix~\ref{ap:details}.

Figure~\ref{tab:H1AndH2} shows data for the object placement for Task 2 (5 participants each feedback type, 15 none). $p$ values for $t$ tests given where $p < 0.1$.

As hypothesized (H1), both the physical and AR interfaces were trusted more than the map interface (first three rows of Table~\ref{tab:H1AndH2}) However, the AR interface was trusted slightly more than the physical one (3rd row). When we look at just the trials where matching feedback was provided  (rows 4-6), this same pattern holds, but not as strongly. This is largely because showing feedback (H2) tended to {\em reduce} overall trust in the indirect measures. This can be seen in rows 7-9, where the majority of the indirect measures indicated {\em less} trust when feedback was provided.

The {\bf survey data} ranked the physical interface with feedback the highest in terms of trust (4.5 out of 5) but the physical interface without feedback and the map interface with feedback the lowest (both 4.1), with AR in between (F - 4.3 and NF - 4.2). Overall, when asked if having/not having feedback would have changed their answer, the participants ranged from A Moderate Amount (2.7 Physical) to A Little (1.7 Map) with AR in between (1.9). Survey trust results improved with feedback for Physical (4.1 to 4.5) and AR (4.2 to 4.3), but declined for Map (4.4 to 4.1). 

The indirect measures, in contrast, tended to {\em decline} when feedback was added, counter to H2. They also indicate that the AR interface was trusted as much (or more) than the Physical one (counter to H1). 

Both the table distance (T1, T3.2) and area measures (A) showed a pattern of Map trusted least (41/32cm, 19,200cm$^2$), then Physical (40/29cm, 11,900cm$^2$), then AR (32/22cm, 11,500cm$^2$). AR was more trusted with feedback (22/20cm, 10,800cm$^2$ vs 42/23cm, 12,200cm$^2$), while Physical feedback reduced trust (42/34cm, 14,500cm$^2$ vs 40/24cm, 9,299cm$^2$). Map was mixed (34/35cm, 19,600cm$^2$ vs 45/29cm, 18,700cm$^2$). 

As seen in Figure~\ref{tab:H1AndH2} (right), the distances of the objects to the edges of the tables in Task 2 also suggest that AR is most trusted (O1, 2, 3, 4), and Map least trusted (O1, 2, 4). We found a significant difference between AR and Map for O1 ($p$=0.046), Physical and AR for O3 ($p$=0.021), and AR and no feedback paired with any interface for O3 ($p$=0.057).

When participants were allowed to choose their own interface plus feedback (Task 3, part 2), all three interfaces showed an increase in trust from the previous tasks (AR 22cm, 11,500cm$^2$ to 20cm, 10,921cm$^2$, Map 32cm, 19,100cm$^2$ to 33cm, 14,300cm$^2$, Physical 29cm, 11,900cm$^2$ to 21cm, 9,400cm$^2$). This could be because the participants were allowed to choose, or because they could validate with a {\em different} feedback.

\noindent{\bf Indirect measure validation:} We compare the object placements and distances for the two fragile versus the two durable objects. The fragile objects were placed on the table close to the edge (trusted) 9/30 and 7/30 times, whereas for the durable objects  it was 14/30 and 15/30 times. The distance results for the objects placed on the close table were mixed (13 \& 15cm versus 15 \& 14cm) but for the far table the fragile objects were placed further from the edge (18 \& 16cm versus 14 \& 15 cm). In summary, the indirect measures did capture differences in trust between the two object categories.

\subsection{H3 \& H4: Interface/Feedback Preference}

The AR interface was the most highly preferred (60\%), followed by the map interface (23\%) and the physical interface (17\%).  This partially supports H3.  

Although the majority of participants found each interface useful (87\%, 90\%, and 97\% for the physical, map, and AR interfaces, respectively), 57\% found the AR interface extremely useful, compared to 43\% for the map interface and 40\% for the physical interface. 

In general, the majority of participants also found each interface to be easy to use (100\% for the physical interface, 93\% for the map interface, and 97\% for the AR interface). However, 73\% of participants found the physical interface extremely easy to use, compared to only 23\% and 53\% for the map and AR interfaces, respectively. 

As hypothesized (H4), the physical feedback was the most highly preferred (73\%), followed by the AR feedback (23\%) and the map feedback (3\%).  When allowed to choose which interface and feedback they wanted to use for the final task, participants chose the AR interface paired with the physical feedback most frequently (40\%) followed by the Map interface with physical feedback (23\%).  

%% file: Text/ResultsFigures/H3.tex
\begin{figure*} [ht]
    \includegraphics[height=0.19\linewidth]{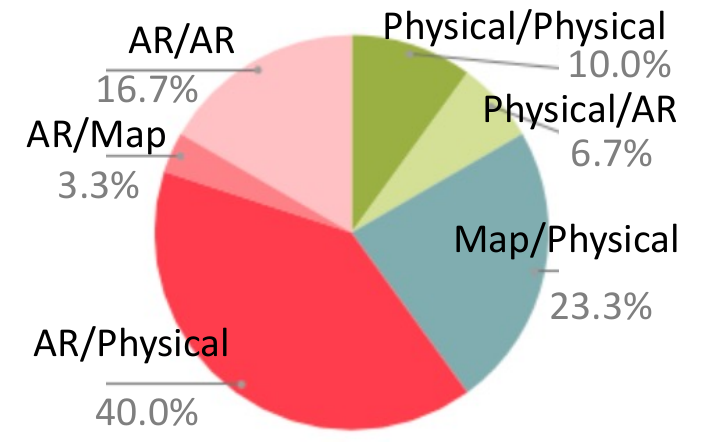}
    \includegraphics[height=0.19\linewidth]{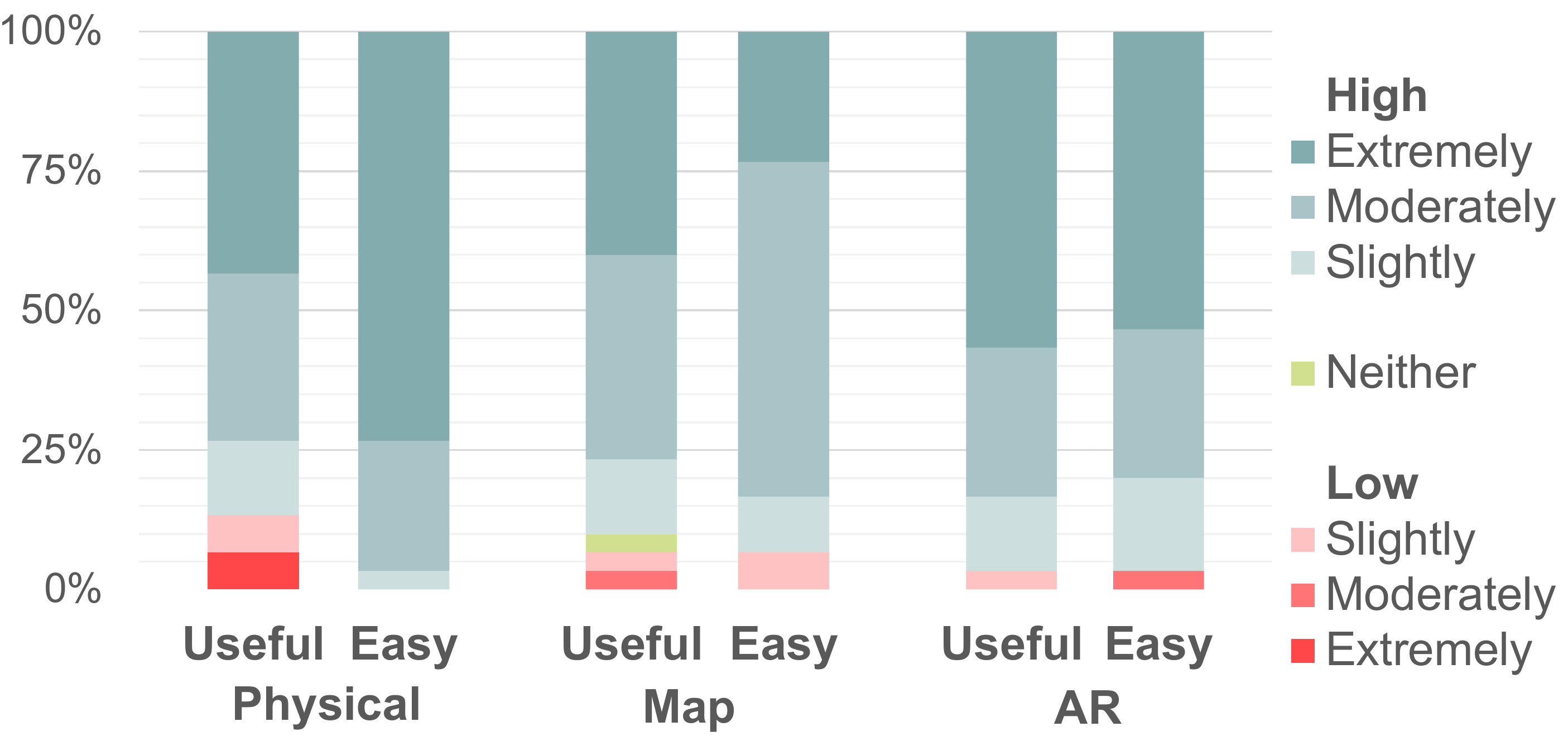}
    \includegraphics[height=0.19\linewidth]{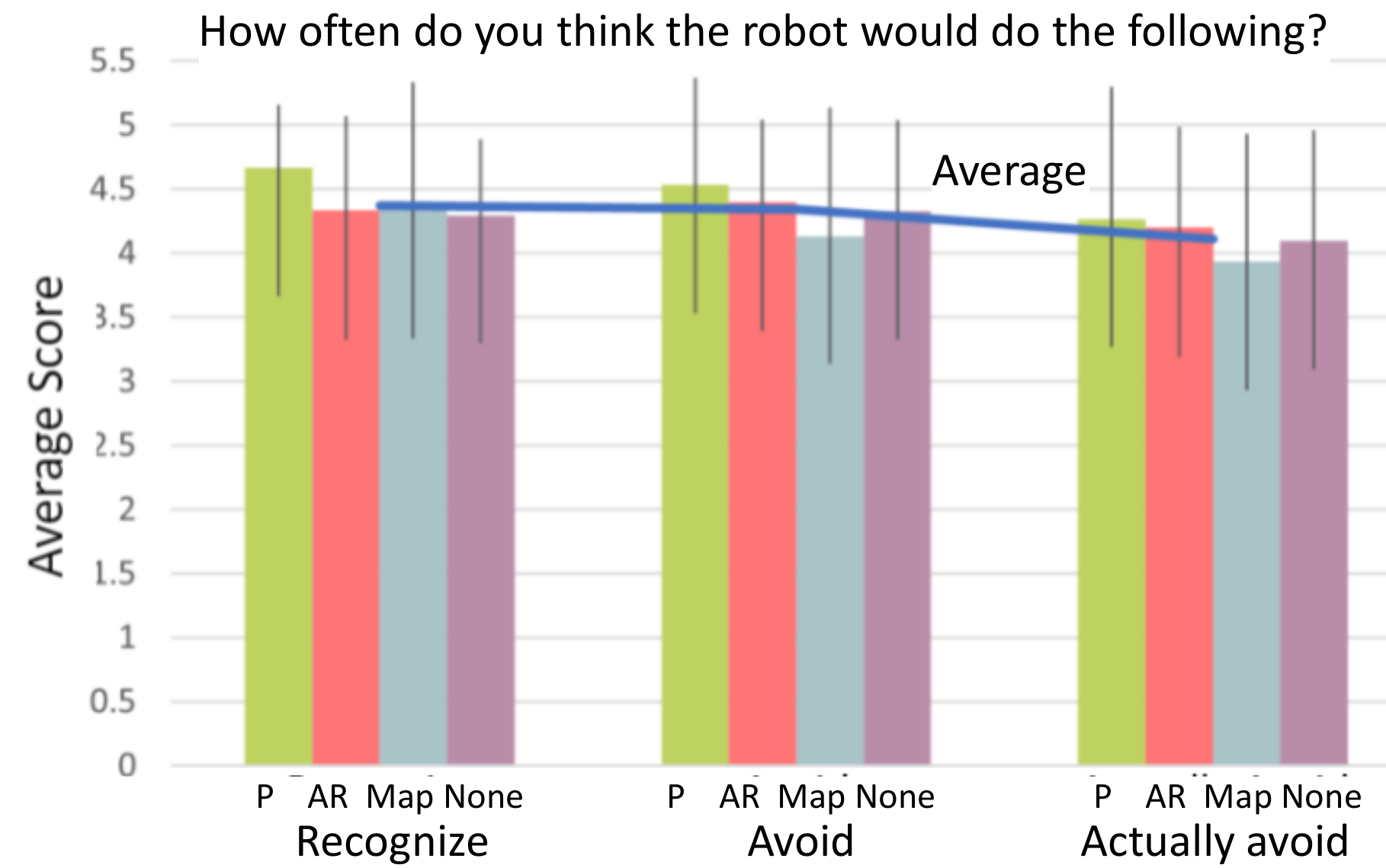}
\caption{Interface/Feedback Preference \textbf{(left)} Usefulness and Ease of Use \textbf{(middle)} Trust survey results \textbf{(right)}.}
\label{fig:InterfacePrefSurvey}
\end{figure*}

%% file: Text/5.discussion.tex
\section{Discussion}
\label{sec:discussion}

The {\bf indirect} measures were internally consistent, but did not always agree with the {\bf direct} measures, in particular for the Physical versus AR comparison and the feedback/no-feedback one. A possible explanation for this is that participant's expectations for the robot are higher than their actual capabilities; when confronted with actual feedback/activities, they unconsciously adjust downwards. This is also reflected in our questions about capabilities, where trust for ``can do'' is lower than ``recognize'' (Figure \ref{tab:H1AndH2} right).

Our most surprising finding is that many participants wanted to use a different feedback from their interface. The popularity of the AR interface (75\%) likely reflects a balancing of familiarity of touch-screen apps with the desirability of the physical interface and feedback (or simply that AR, being novel, is more ``cool''). The mix may also be a form of ``data triangulation'', where the alternate feedback measures the ``whole'' system, not just the interface.

In summary, different interface/feedback combinations {\em do} influence trust and self-reported measures of trust may be higher than actual trust, particularly when the participant sees the robot's actual behavior. Not surprisingly, AR interfaces are a good mix of familiar/easy to use while maintaining the benefits of operating ``in the physical world.'' 

\textbf{Limitations:} 
    The study took place in a staged environment.  It had the look and feel of an office lounge area, but the user did not own the furniture, objects, or the robot.  We cannot be certain that our results would extend to a person's real living room or office.
    We are also only measuring trust and preference in the first encounter.  There is the potential that a user's level of trust in the robot or interface preference may change over time.   

%% file: Text/7.conclusions.tex
\section{Conclusions}
We conducted a user study to examine the influence interfaces and feedback types have on human-robot trust when specifying a no-go zone for a mobile robot. We found that the choice of interface and feedback does influence levels of trust and that participants preferred to use a different modality for the interface than the feedback.  In particular, participants preferred to use the augmented reality interface to specify a no-go region and then to receive physical feedback in the form of the robot driving around the marked off area.

%% file: Text/9.appendix.tex
\begin{table*}
 \centering
 \begin{tabular}{|l|ccc|}
 \hline 
 Comparison       &  Support & Equiv & Not support \\ 
 \hline 
 Phys I $>$ Map I & O1 O2 D2 T3.1 A & T1 O34 D1 D3 D4 S1-3 &   \\
 AR I       $>$ Map I & T1 O1 O3 O4 T3.1 A D1-4 & O2 S1-3 & \\
 Phys I $>$ AR I  & O1 A & O2 O4 D1 D2 D4 S1-3 & T1 O3 D3 T3.1 \\
 \hline 
 Phys IF $>$ Map IF & O1 O4 D1 D4 S1-3 T3.1 A & O2 D2 & T1 O3 D3\\
 AR IF       $>$ Map IF & T1 O134 D134 S23 T3.1 A & O2 D2 S1 & \\
 Phys IF $>$ AR IF & O1 S1 S2 & O2 O4 D1 D2 A & T1 O3 D3 D4 S3 T3.1\\
\hline
 Phys F $>$ No F & O4 S1 S2  S3 & D1 D2 D3 A & T1 O1 O2 O3 D4 T3.1 \\
 AR F       $>$ No F & T1 D1 D3 S2 S3 A T3.1 & O2 O4 D2 D4 S1 & O1 O3 \\
 Map F $>$ No F & T1 O3 D2 S4 & O2 & O1 O4 D3 D4 S1 S2 S3 T3.1 A\\
\hline 
Feedback $>$ No F & S1 & O4 D1-4 S2 S3 & O1-3 A \\
\hline
 \end{tabular}
 \caption{Specifics of trust measures for H1 and H2.}
 \label{tab:H1AndH2Details}
 \end{table*}

\appendix
\section{Trust measure details}
\label{ap:details}

Table~\ref{tab:H1AndH2} summarizes the metrics in support of specific comparisons both for the interface (I) and the feedback used (F). The metrics are: Placement of the table within the region (Task 1-T1 and Task 3-T3.1, T3.2), object on the close table versus the far (Task 2, O1-4), distance of the object from the edge of the table (Task 2, D1-4), area marked out for the no-go region (Task 3, part 2, A), survey questions (Tasks 1, 2, and 3.1, S1-3), and whether or not their trust would have changed with the feedback (Tasks 1, 2, and 3.1, C). Table rows are as given in Table~\ref{tab:H1AndH2}.